\documentclass[pmlr]{jmlr}

\usepackage{algorithm}
\usepackage{amsmath}
\usepackage{enumerate}
\usepackage{mdframed}
\usepackage{enumitem}
\usepackage{booktabs}
\usepackage{caption}
\usepackage{multirow}
\usepackage{hyperref}
\usepackage{natbib}

\usepackage{longtable}

 \usepackage{booktabs}
 
\usepackage[load-configurations=version-1]{siunitx} 

\makeatletter
\def\set@curr@file#1{\def\@curr@file{#1}} 
\makeatother

\jmlrvolume{}
\jmlryear{2023}
\jmlrworkshop{To appear at Machine Learning for Healthcare 2023}

\title[PCBFL]{Privacy-preserving patient clustering for personalized federated learning}

\author{\Name{Ahmed Elhussein}
       \Email{ae2722@cumc.columbia.edu}\\ 
       \addr Department of Biomedical Informatics, Columbia University\\
       New York City, NY, U.S.A
       \AND
       \Name{Gamze G{\"u}rsoy}
       \Email{gamze.gursoy@columbia.edu}\\ 
       \addr Department of Biomedical Informatics, Columbia University\\
       New York Genome Center\\
       New York City, NY, U.S.A} 

\begin{document}

\maketitle

\begin{abstract}
Federated Learning (FL) is a machine learning framework that enables multiple organizations to train a model without sharing their data with a central server. However, it experiences significant performance degradation if the data is non-identically independently distributed (non-IID). This is a problem in medical settings, where variations in the patient population contribute significantly to distribution differences across hospitals. Personalized FL addresses this issue by accounting for site-specific distribution differences. Clustered FL, a Personalized FL variant, was used  to address this problem by clustering patients into groups across hospitals and training separate models on each group. However, privacy concerns remained as a challenge as the clustering process requires exchange of patient-level information. This was previously solved by forming clusters using aggregated data, which led to inaccurate groups and performance degradation. In this study, we propose Privacy-preserving Community-Based Federated machine Learning (PCBFL), a novel Clustered FL framework that can cluster patients using patient-level data while protecting privacy. PCBFL uses Secure Multiparty Computation, a cryptographic technique, to securely calculate patient-level similarity scores across hospitals. We then evaluate PCBFL by training a federated mortality prediction model using 20 sites from the eICU dataset. We compare the performance gain from PCBFL against traditional and existing Clustered FL frameworks.  Our results show that PCBFL successfully forms clinically meaningful cohorts of low, medium, and high-risk patients. PCBFL outperforms traditional and existing Clustered FL frameworks with an average AUC improvement of 4.3\% and AUPRC improvement of 7.8\%.
\end{abstract}

\section{Introduction}

The use of deep learning on Electronic Health Records (EHR) has been widely and successfully implemented for a range of goals such as for disease risk prediction, diagnostic support, and Natural Language Processing (\cite{Esteva2019-ex, Gulshan2016-yh, Miotto2016-hl, Choi2016-ly}). However, to leverage the predictive ability of deep learning models on the inherently high dimensionality of EHR data, a large number of samples are needed. Undersampled or overspecified models are more likely to overfit on training datasets and generalize poorly when applied to new datasets (\cite{hosseini2020tried, Miotto2018-nt}). This is especially important in rare disease settings where a single institution cannot have enough power to develop predictive models. One solution to enable more sophisticated and accurate models is to increase available training data. While some attempts to build large and diverse cohorts have been successful (\textit{e.g.,} All Of Us and UK Biobank), they are largely volunteer based and limited in the number and diversity of patients enrolled, with the majority of patients coming from healthy populations.  Another alternative is institutional data-sharing but the regulatory framework (\textit{e.g.,} HIPAA) and the ethical need to respect patient privacy limit widespread data sharing across institutions. One solution to support collaborative learning across sites while minimizing privacy concerns is Federated learning (FL) (\cite{McMahan2017-qm, kaissis2020secure}). FL is a distributed machine learning approach  that enables multiple sites to collaboratively train a model while keeping data local. The process involves sites sharing locally trained model parameters with a central server, which then aggregates these parameters to create a global model. This process is repeated for a number of training rounds until a final global model is obtained. The parameters are aggregated via a commonly used algorithm, Federated Averaging (FedAvg), which uses sample-size weighted averaging to combine model parameters. FL enables model training on larger and more diverse patient groups across many sites while keeping datasets local. Furthermore, it has the benefit of allowing sites with limited training data  (\textit{e.g.,} rural hospitals) to be involved in model building. That is, FL has the potential to improve model performance, generalizability, and fairness (\cite{Rieke2020-ck}).  As such, FL has become increasingly popular in healthcare, and has been implemented on a range of tasks including disease risk prediction, diagnosis, and image recognition (\cite{Rieke2020-ck, Dayan2021-ze, Pati2022-wv}). 

FL still has some limitations. FedAvg underperforms when data is non-identically independently distributed (non-IID) across sites. This is a particular concern for EHRs, where a range of factors can lead to distribution shifts including patient composition, institutional treatment guidelines, and institutional data capture processes (\cite{zhao2018federated}). Patient composition, \textit{i.e.,} differences in demographics and clinical presentation is one of the most significant sources of distribution shift (\cite{Prayitno2021-gm}). Personalized FL, which aims to account for distribution shifts across datasets, is a potential solution for training models on non-IID data (\cite{fallah2020personalized}). Clustered Federated Learning is a variant of personalized FL that has demonstrated success in handling non-IID data when datasets naturally partition into clusters (\textit{e.g}., clinical groups) (\cite{Ghosh2020-wp}). In this scenario, training separate models for each cluster has been shown to improve performance on downstream tasks. The challenge lies in identifying the clusters and partitioning the datasets accordingly. Recent patient clustering preprocessing using individual patient embeddings demonstrate improvements in downstream task performance in the centralized setting (\cite{Xu2020-ky, Zeng2021-yh}). This suggests clustering can be a promising avenue for healthcare tasks in a federated setting as well. 
 
Addressing the absence of a privacy-preserving federated approach for clustering using individual patient embeddings is a critical step for personalized FL. Frameworks such as Differential Privacy (DP) have been introduced to address this issue. DP works by adding noise to summary-level data (\textit{e.g.,} model parameters) prior to sharing information with a central server. But in the clinical context, the amount of noise needed to achieve privacy can compromise model performance (\cite{Ficek2021-vt, Dwork2014-qi}). One can use cryptographic techniques that provide mathematical privacy guarantees without adding noise and can work with both summary and individual level data. One such technique that is appropriate in a multi-site setting is Secure MultiParty Computation (SMPC). SMPC enables multiple parties to jointly compute a function over their inputs while keeping those inputs secret from each other by using a secret-sharing scheme (\cite{Evans2018-oj}). 

In this study, we introduce Privacy-preserving Community-Based Federated machine Learning (PCBFL), a privacy-preserving framework that incorporates a clustering preprocessing step into FL (Clustered FL). Using SMPC, PCBFL securely calculates patient-level embedding similarities across all sites while preserving privacy. We assume an honest-but-curious adversary scenario, in which the computing parties cannot learn the input from the secrets and will not intentionally collude with each other to learn the input (\cite{Evans2018-oj}). By using individual patient embedding similarity scores to cluster patients into groups, we aim to improve downstream task performance. We evaluate PCBFL algorithm against two main federated comparators on a downstream mortality prediction task: Community-Based Federated machine Learning (CBFL) and FedAvg. CBFL, a state-of-the-art method, also employs a clustering preprocessing step, however, unlike PBCFL, CBFL uses aggregate hospital embeddings for patient clustering. FedAvg is the standard algorithm with no preprocessing for non-IID data. Additionally, we assess the performance of non-federated algorithms that conduct only model training: single site training and centralized training. Single site performance serves as a baseline that all federated algorithms should surpass, while centralized performance represents the gold standard against which federated algorithms are compared. We show that our PCBFL approach results in improved performance compared to both standard FedAvg and CBFL on the mortality prediction task. PCBFL also outperforms FedAvg and CBFL in the majority of the individual sites. In addition, our results demonstrate that PCBFL produces clinically meaningful clusters, grouping patients in low, medium, and high-risk cohorts. This suggests that PCBFL has the potential to support other clustering tasks such as federated phenotyping. Future work could explore the utility of our approach in a wider range of clinical applications. 

\subsection{Generalizable Insights about Machine Learning in the Context of Healthcare}
In healthcare, protecting patient privacy while leveraging data to improve clinical outcomes is a crucial challenge. This challenge is particularly relevant for complex deep learning models that require large sample sizes. Our paper introduces PCBFL, a privacy-preserving framework that uses SMPC to incorporate a clustering preprocessing step into federated learning. Our approach provides a practical solution to protect patient privacy while enabling patient-level calculations across different hospitals. Another key issue in healthcare is dealing with non-IID data, where data from different sites have different distributions. Our clustering-based approach effectively handles non-IID data by partitioning patients into clinically relevant groups. This clustering procedure can  be used in various healthcare tasks, including unsupervised patient clustering for phenotyping. Our methodological contributions could be extended to conduct large-scale phenotyping across many sites, enabling more accurate and granular sub-phentoypes to be discovered.

\section{Related work}
In FL, several works have studied the statistical heterogeneity of users’ data and linked high heterogeneity to performance degradation and poor convergence (\cite{li2018convergence}). To address this, researchers have attempted to personalize learning to each user (\cite{tan2022towards}). The proposed solutions typically occur at the preprocessing, learning, or postprocessing stages. Preprocessing solutions include data-augmentation and client partitioning (\cite{sattler2020clustered, zhao2018federated}). Learning solutions include meta-learning and modifications to the FedAvg algorithm (\textit{e.g.,} addition of regularization parameters) (\cite{fallah2020personalized, deng2020adaptive, li2020federated1}). Post-processing techniques involve adaptation of the global model by the local site after federated training is complete (\cite{hanzely2020federated}). In healthcare, personalized FL has mostly focused on preprocessing steps. An example is CBFL, which uses embeddings to cluster patients (\cite{Huang2019-pm}). The authors showed clustering improved performance on a downstream mortality prediction task compared to the standard FedAvg technique. However, due to privacy constraints, CBFL’s patient clustering is based on average embeddings per site, \textit{i.e.,} it does not use individual patient embeddings. This led to patient clusters based on the geography of hospitals and not based on patient characteristics.

\section{Methods}
\subsection{Cohort and feature extraction}
We used the eICU collaborative research database, which contains critical care data for 200,859 patients at 208 hospitals across the United States (\cite{Pollard2018-hw}). We followed a similar data-processing step as \cite{Huang2019-pm}. The outcome of interest was mortality in the ICU, defined as the unit discharge status (0 for alive and 1 for expired). The independent variables are diagnosis, drugs, and physical exam markers in the first 48 hours of admission. We limited our features to the first 48 hours to ensure consistency on patient follow-up times and clinical relevancy of model predictions. 

For diagnosis and drugs, we used the count of times the feature appears in the dataset for that patient. For physiological markers, we used the first recorded instance. Physical exam markers used include: Glasgow Coma Scale (GCS) Motor, GCS Verbal, GCS Eye, Heart Rate (HR), Systolic Blood pressure (SBP), Respiratory rate (RR) and Oxygen Saturation (O2\%), age, admission weight, admission height. Drug and physiologic features were kept as is (1,056 and 7 features in total, respectively). Diagnosis codes were rolled up to 4 digits, \textit{i.e.,} all 5 digit codes were converted to 4 digit codes resulting in 483 diagnosis codes. Note, compared to \cite{Huang2019-pm}, we also added diagnosis and physical exam markers to the features as these have been shown to improve predictive performance in related tasks (\cite{Sheikhalishahi2020-bi}). All data was 0-1 normalized prior to training by the models. 

We extracted patients from the dataset who had data for all three variable groups and a recorded outcome. This was done to avoid the need for imputation which could introduce bias to the data. Doing so would affect evaluation of the training architectures. We then filtered for sites that had a minimum of 250 patients, resulting in 20 sites and 20,221 patients. We randomly subsample 250 patients from each site, creating a final cohort of 5,000 patients. This subsampling approach was intended to create a more realistic FL scenario, where size of the dataset in each site is limited. 

\subsection{Privacy-preserving CBFL} 
A schematic of the steps involved in PCBFL is presented in Figure 1 and the procedures for each step are detailed in Supplementary Algorithms 1-4 (Appendix E). PCBFL is  composed of four procedures: creating patient embeddings, estimating patient similarity securely, clustering patients, and predicting mortality.

\begin{figure}[!ht]
  \centering 
  \includegraphics[width=9cm, height=10cm]{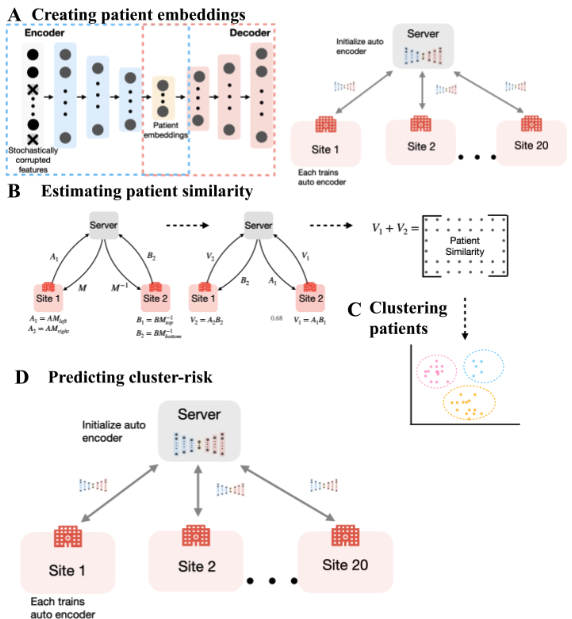} 
\vspace*{-4mm}
  \caption{\textbf{(A) Training a denoising autoencoder to create embeddings.} A federated autoencoder is trained to obtain latent variables for each feature domain. Latent variables are concatenated to form a patient embedding vector. \textbf{(B) SMPC protocol to calculate the cosine similarity between vectors.} SMPC uses a secret sharing scheme to jointly calculate the dot product between pairs of vectors. \textbf{(C) Spectral clustering to cluster the patients} using similarity matrix generated from pairwise cosine similarities of embeddings. \textbf{(D) Cluster-based FL training.} Each model is separately trained per cluster.
}
  \label{fig:example} 
\end{figure}

\subsubsection{Creating patient embedding} 
Following \cite{Huang2019-pm}, we trained a federated denoising autoencoder made up of 6 layers including a three-layer encoder and an identical three-layer decoder to create patient embeddings. To reduce overfitting, 30\% of the features are stochastically corrupted during training, \textit{i.e.,} 30\% features are forced to 0. A separate autoencoder was trained for each feature domain, \textit{i.e.,} drugs, diagnosis, and physical examination. We used a ReLU activation in the hidden layers, a sigmoid activation in the final output layer, and a Mean Squared Error loss. We used an Adam optimizer with a learning rate of 1$e^{-3}$ and batch size of 32. Federated models were trained for 20 rounds with 10 epochs per round and centralized models were trained for 200 epochs. For one patient’s embedding, we concatenated the latent variables of each feature domain

\subsubsection{Estimating patient similarity securely} 
We used cosine similarity as the similarity metric as it is invariant to scaling effects and works well with high dimensional vectors compared to euclidean distance (\cite{Strehl2000-ju, Li2022-sa}). We used SMPC to securely calculate  patient embedding similarity across sites while preserving privacy. SMPC is a cryptographic technique that allows parties to jointly compute a function over their inputs while keeping the inputs secret, \textit{i.e.,} only the output is made available (\cite{Evans2018-oj}). The benefit of SMPC is that it protects privacy against both outside adversaries and other involved parties with mathematical guarantees and allows for exact calculation of cosine similarity across sites. We adapted a protocol from \cite{Du2004-pq} to calculate the dot product across sites (see SMPC protocol) using secret sharing for an honest-but-curious adversary model. This protocol involves the following steps: 
\begin{enumerate}
  \item Create a $dxd$ invertible matrix $M$ and send $M$ to $site_1$ and $M^{-1}$ to $site_2$ (where $d$ is the embedding dimension)  
  \item Each site divides their dataset into submatrices and masks them with $M$ or $M^{-1}$ 
  \item A limited number of masked submatrices are shared between sites
  \item The submatrices are combined to produce the final dot product without revealing any information about the dataset
\end{enumerate}

\begin{mdframed}
\section*{SMPC PROTOCOL}
$Site_1$ hold dataset A ($N_1\times d$), $Site_2$ hold dataset B ($N_2\times d$) where $d=$embedding dim and $N_i=$ number of patients.
\begin{enumerate}
\item Server creates a random invertible matrix $M_{d\times d}$ using Reed-Hoffman encoding and sends $M$ to Site 1 and $M^{-1}$ to Site 2.
\item Site 1 computes $A_1 = A\times M_\text{left}$, $A_2 = A\times M_\text{right}$ and sends $A_1$ to the server.
\item Site 2 computes $B_1 = B\times M_\text{top}^{-1}$, $B_2 = B\times M_\text{bottom}^{-1}$ and sends $B_2$ to the server.
\item Server sends $B_2$ to Site 1 and $A_1$ to Site 2.
\item Site 1 computes $V_a = A_2\times B_2$ and sends it to the server.
\item Site 2 computes $V_b = A_1\times B_1$ and sends it to the server.
\end{enumerate}
\end{mdframed}

We have $\mathbf{AB} = AMM^{-1}B = \begin{pmatrix} A_1 & A_2 \end{pmatrix} \begin{pmatrix} B_1\\ B_2 \end{pmatrix} = V_a + V_b$.

Secure calculation is possible as no party has sufficient information to reconstruct the original dataset with only some of the submatrices.  More concretely, it can be seen that as long as sites only share half of their encoded matrices  ($A_1$ and $B_2$) there remains an infinite number of solutions to the problem. The method relies on the construction of a secure matrix $M$. This matrix can be generated using maximum distance separable (MDS) codes such as Reed-Solomon codes\cite{Du2004-pq} . MDS codes ensure that any subset of columns are linearly independent of each other making it impossible to recover the original data. For a more detailed introduction on MDS codes please see \cite{macwilliams1977theory}. Note that all embeddings were first L2-normalized prior to calculating the dot product so this product is equivalent to cosine similarity. In our case, we conduct pairwise calculation of cosine similarity between sites and concatenate these  together to construct the final similarity matrix across all patients.

\subsubsection{Clustering patients}
We employed spectral clustering to cluster patients using the cosine similarity matrix. Spectral clustering is suited to this task as it utilizes the similarity matrix’s global structure to capture complex relationships (\cite{Von_Luxburg2007-eu}). We used the elbow method to determine the optimal number of clusters. For this, we first calculated the Within-Cluster-Sum-of-Squares (WCSS) for clusters 1-10 (see Supplement: Appendix A.1 WCSS). WCSS is a metric to measure the compactness of the clusters. We selected the ‘elbow’ point of the plot after which additional clusters do not lead to substantial improvements in WCSS (\emph{i.e.,} compactness of the clusters). This is a heuristic  that determines the minimum number of clusters necessary to account for the majority of the variance in the dataset (\cite{madhulatha2012overview}). A smaller WCSS implies that the data points are more compact, indicating tighter clustering of similar points. We assessed a range of clusters between 1 to 10 and ultimately choose 3 (Supplementary Figure 1).

\subsubsection{Predicting mortality}
We trained a FeedForward neural network with 3 input heads and a classification module. Each input head processes a feature domain into a 5-dimensional representation. These representations are concatenated and fed through the classification module to generate predictions. The multihead structure integrates distinct data domains more effectively by first processing each type separately before combining them for prediction. We followed a similar training algorithm as FedAvg, but we trained a model on each cluster separately. That is, the server initializes a separate model for each cluster and each site trains a cluster model only on the patients in that respective cluster. Weights were aggregated based on each site's sample size for that cluster (Figure 1D). 
 
\subsection{Other models}
We evaluated the performance gain from PCBFL against other training methods including: single site, centralized, FedAvg and CBFL. Single site training refers to the average performance of each site if it were to train a model separately. Centralized training refers to the performance of a model trained on all data together as if it were one site. Centralized training performance is the gold-standard benchmark hoped to be achieved by FL. FedAvg refers to standard Federated Averaging procedure, which aggregates model parameters based on sample size in each site. CBFL refers to the community based clustering procedure described by \cite{Huang2019-pm} that uses K-means clustering on average hospital embeddings. The generated clusters were sent to each site to assign a cluster to each patient. Models are then trained separately for each cluster. Note that the major difference between CBFL and PCBFL is in the clustering approach; CBFL uses average site embeddings while PCBFL enables privacy-preserving clustering of patients based on individual patient embeddings. 

\subsection{Model training}
We implemented the feedforward models with ReLU activation in the hidden layers and a sigmoid activation in the final output layer. We employed Binary Cross Entropy loss. All feedforward models were trained using the same hyperparameters: an Adam optimizer with a learning rate of 1$e^{-3}$ for all models and batch size of 32. For all federated training methods, we used 20 training rounds with 10 epochs per round. For all central models, we used 200 epochs. This kept the training epochs consistent across all models.

\subsection{Evaluation}
\subsubsection{Cohort analysis for clustered FL algorithms}
We examined the clusters generated by PCBFL and CBFL by comparing patients’ mortality and feature distributions between clusters. We used one-way ANOVA testing for continuous variables and Negative Binomial testing for count variables to determine statistical significance. We chose Negative Binomial over Poisson to account for the overdispersion in the count data (\cite{Hilbe2011-qu}). A p value of $<$0.05 with Bonferroni correction was used to determine statistical significant differences between clusters. We also examined whether the clusters generated by PCBFL and CBFL capture the regional distribution of hospitals. We grouped hospitals by region as defined by eICU Collaborative Research Dataset (Midwest, Northeast, South, West) and conducted a chi-squared test examining the relationship between region and cluster distribution.

\subsubsection{Prediction task}
Data was randomly split into training and testing datasets in a 70:30 ratio. Since only 20\% of the labels were positive, we evaluated performance with both AUC and AUPRC scores. We ran the models for 100 times and calculated the mean scores and bootstrapped estimates (1000 iterations) of the 95\% confidence intervals. We calculated the overall performance of a protocol as the weighted average of individual site scores. This weighting accounts for the number of samples used in model development at each site. For the cases of Single Site and FedAvg, where each site trains one model and has the same number of patients, this is a simple average. In the cases of CBFL and PCBFL, where sites train three separate models, the weighting is based on the proportion of patients that belong to the site and cluster (see Supplement: Appendix B. Calculating overall performance). We also compared the performance of the models at each site to determine if there are sites that fail to benefit from FL, or PCBFL in particular. This is important as we wanted to  ensure that all sites benefit from FL to incentivize FL collaboration \cite{li2020federated1}. All code was written in Python 3.9.7 and Pytorch 1.12.1 and is available on GitHub https://github.com/G2Lab/pcfbl.

\section{Results} 
\subsection{PCBFL provides privacy-preserving and accurate patient similarity scores using Secure MultiParty Computation}
We first evaluated whether the use of SMPC affects the accuracy of the patient similarity scores calculated using cosine similarity. This was done by comparing PCBFL results to the results of a plaintext and centralized calculation (referred to as True). The root mean squared error between the True and privacy-preserving scores was $<$5$x10^{-11}$, indicating that the protocol is highly accurate. Figure 2 displays the comparison of the True cosine similarity scores and privacy-preserving cosine similarity scores, where each point represents a pairwise comparison of two patients. The graph demonstrates that the SMPC protocol accurately calculates the cosine similarity between patients' embeddings at all ranges.
\begin{figure}[!ht]
  \centering 
  \includegraphics[width=10cm, height=7cm]{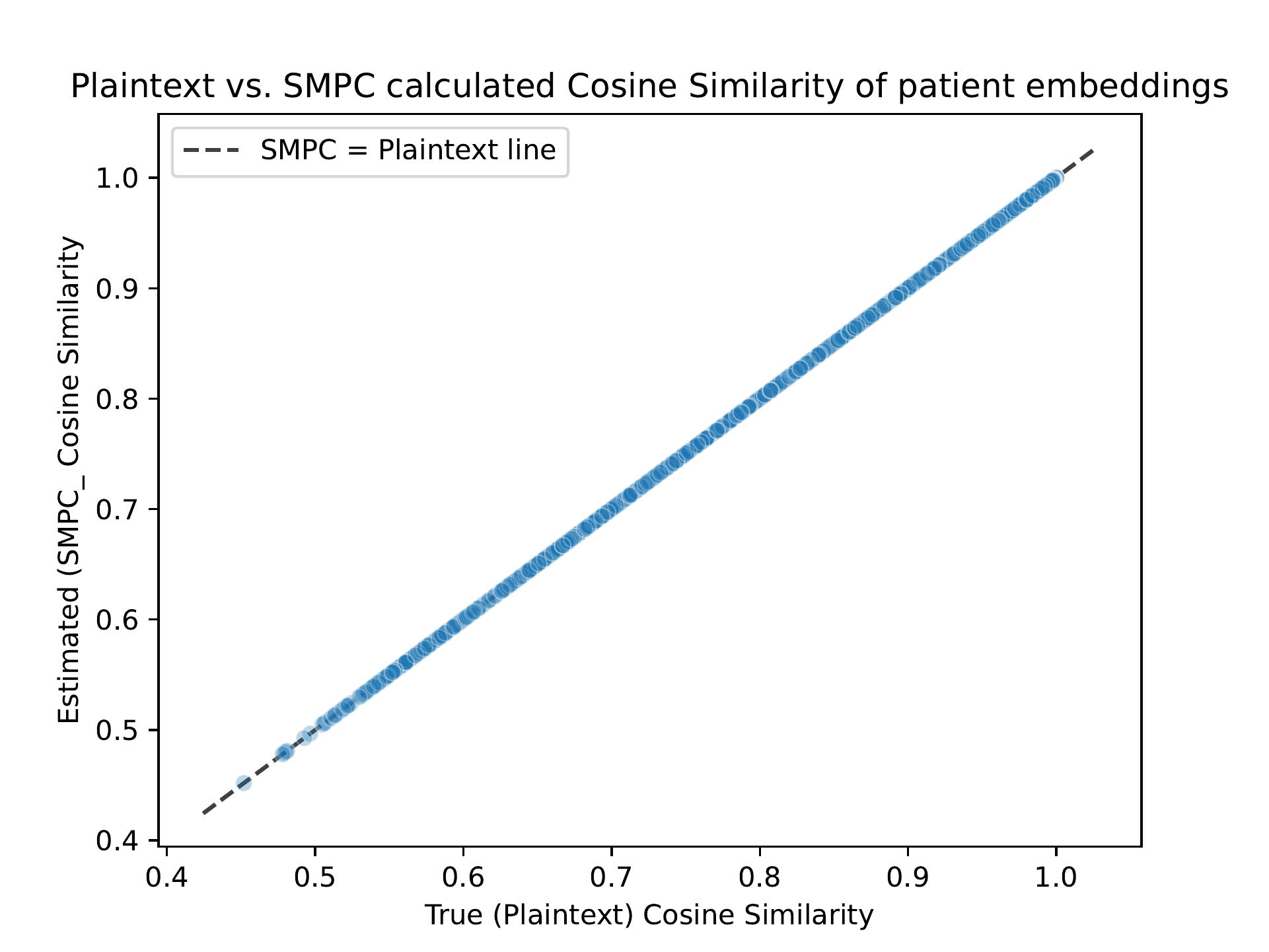} 
  \vspace*{-4mm}
  \caption{\textbf{Comparison of cosine similarity scores} calculated using PCBFL’s SMPC protocol and a plaintext centralized truth (True). Each point is a cosine similarity score between 2 patient embeddings.
}
  \label{fig:example} 
\end{figure} 
\vspace*{-5mm}

\subsection{PCBFL provides clinically meaningful clusters} 
We examined the clusters determined by PCBFL in terms of their mortality and physical examination scores (Table 1). The resulting three clusters were found to correspond to three distinct levels of severity based on both true mortality rates and physical examination scores.The high risk group was more likely to have higher mortality, lower GCS, higher age and worse vital sign measurements (p $<$ 0.005). In contrast, an examination of CBFL did not yield distinct clinical severity groups, with only significant differences in age and RR. Full mortality and physical examination feature distributions of PCBFL and CBFL clusters are shown in Supplementary Tables 2 and 3, respectively (Appendix F). 

We also compared the distribution of diagnosis counts across clusters and found that conditions indicative of severe disease are more likely to occur in the high-risk group (Table 2, see Methods 3.5.1 for statistical tests used). Overall, we identified 28 out of 483 diagnoses that were more likely to occur in the high risk group  (p$<$0.0001). These included clinically relevant diagnoses for cardiovascular disease, respiratory disease, renal disease, infectious disease, metabolic disorders, hematological disorders, and nutritional disorders. A similar analysis on CBFL clusters yielded no statistically significant differences in diagnosis counts. Full diagnosis feature distributions of PCBFL and CBFL clusters are shown in Supplementary Tables 4 and 5, respectively (Appendix F). 

\begin{table}[t]
  \centering 
  \caption{Outcome and physical examination distribution for 3 clusters identified by PCBFL}
  \begin{tabular}{lllll}
  \toprule
    \textbf{PCBFL clusters:} & \textbf{Low} & \textbf{Medium} & \textbf{High} & \textbf{p-value} \\
    \midrule
    \textbf{Mortality} & 11.7\% & 20.8\% & 26.3\% & $<$0.005* \\ 
    \textbf{GCS} & 13.1 & 12.7 & 12.5 & $<$0.005* \\ 
    \textbf{Age} & 55.4 & 67.8 & 69.2 & $<$0.005* \\ 
    \textbf{HR} & 86.7 & 89.1 & 92.6 & $<$0.005* \\ 
    \textbf{SBP} & 127.0 & 118.0 & 116.1 & $<$0.005* \\ 
    \textbf{RR} & 20.2 & 20.1 & 21.4 & $<$0.005* \\ 
    \textbf{O$_2$\%} & 97.2 & 96.3 & 96.1 & $<$0.005* \\ 
    
    \bottomrule
  \end{tabular}
  \label{tab:example}  \\
  * statistically significant
\end{table}

We performed the same comparison for the  distribution of medication counts between clusters and found 154 drugs out of 1,056 that differed between clusters. Specifically,  45 and 109 in drugs were more likely to be prescribed in the high and medium risk groups, respectively (p$<$5$e^{-5}$). Same analysis for CBFL clusters resulted in 48 out of 1,056 drugs that differed between clusters. Full medication count distributions of PCBFL and CBFL clusters are shown in Supplementary Tables 6 and 7, respectively (Appendix F). Finally, comparing the distribution of clusters by region showed that PCBFL clusters are not associated with region (p=0.10) but CBFL clusters are (p<1$e^{-3}$). See Supplementary Table 1 (Appendix C) for the full cluster distribution breakdown by region. 

\begin{table}[t]
  \centering 
  \caption{Conditions more likely to occur in PCBFL high-risk group}
  \begin{tabular}{ll}
  \toprule
    \textbf{Category} & \textbf{Condition more likely to occur (p $<$ 0.0001)}  \\
    \midrule
    \textbf{Cardiovascular} & Atrial Fibrillation and Flutter, Congestive Heart Failure,\\ &  Hypertension, Tachycardia \\ 
    \textbf{Respiratory} & Asphyxia and Hypoxemia, Obstructive Chronic Bronchitis, \\ & Paralysis of Vocal Cords or Larynx\\ 
    \textbf{Renal} & Acute Kidney Failure, Chronic Kidney Disease, Cystitis \\ 
    \textbf{Infectious} & Septicemia, Fever \\ 
    \textbf{Metabolic} & Abnormal Blood Chemistry, Acidosis, Disorders of Magnesium Metabolism, \\& Hyperlipidemia, Hyperpotassemia, Hypothyroidism, Obesity \\ 
    \textbf{Hematologic} & Anemia, Coagulation Defects, Disease of White Blood Cells,\\ & Thrombocytopenia \\ 
    \textbf{Nutritional} & Protein-calorie Malnutrition, Dehydration \\ 
    \bottomrule
  \end{tabular}
  \label{tab:example} 
\end{table}

\subsection{PCBL increases predictive performance of federated learning } 
Next, we evaluated performance on the mortality prediction task. Table 3 shows the global AUC and AUPRC scores of model training for Single site, Centralized, FedAvg, CBFL and PCBFL. PCBFL achieves statistically significant improvements against Single site, FedAvg and CBFL. Compared to CBFL and FedAvg, PCBFL improves mean AUC by 4.4\% (3.0-5.5\% at 95\% CI) and 4.2\% (2.8-5.8\% at 95\% CI)  and AUPRC by 7.3\% (3.4-11.6\% at 95\% CI) and 8.4\% (3.4-13.8\% at 95\% CI) , respectively. Figures 3a and 3b show global AUC and AUPRC scores for each model. Note that we calculated average per site performance for FedAvg and Single site training, while performance was measured as a weighted average of per cluster and site performance for CBFL and PCBFL (see Supplement Appendix B for definitions).

\begin{figure}[!ht]
\floatconts
{fig:example2}
{\vspace*{-8mm}\caption{\textbf{Performance by model type}, AUC (a) and AUPRC (b) for Single, Centralized, FedAvg, CBFL and PCBFL.}}
{%
\subfigure{%
\label{fig:pic1}
\includegraphics[width=0.49\textwidth]{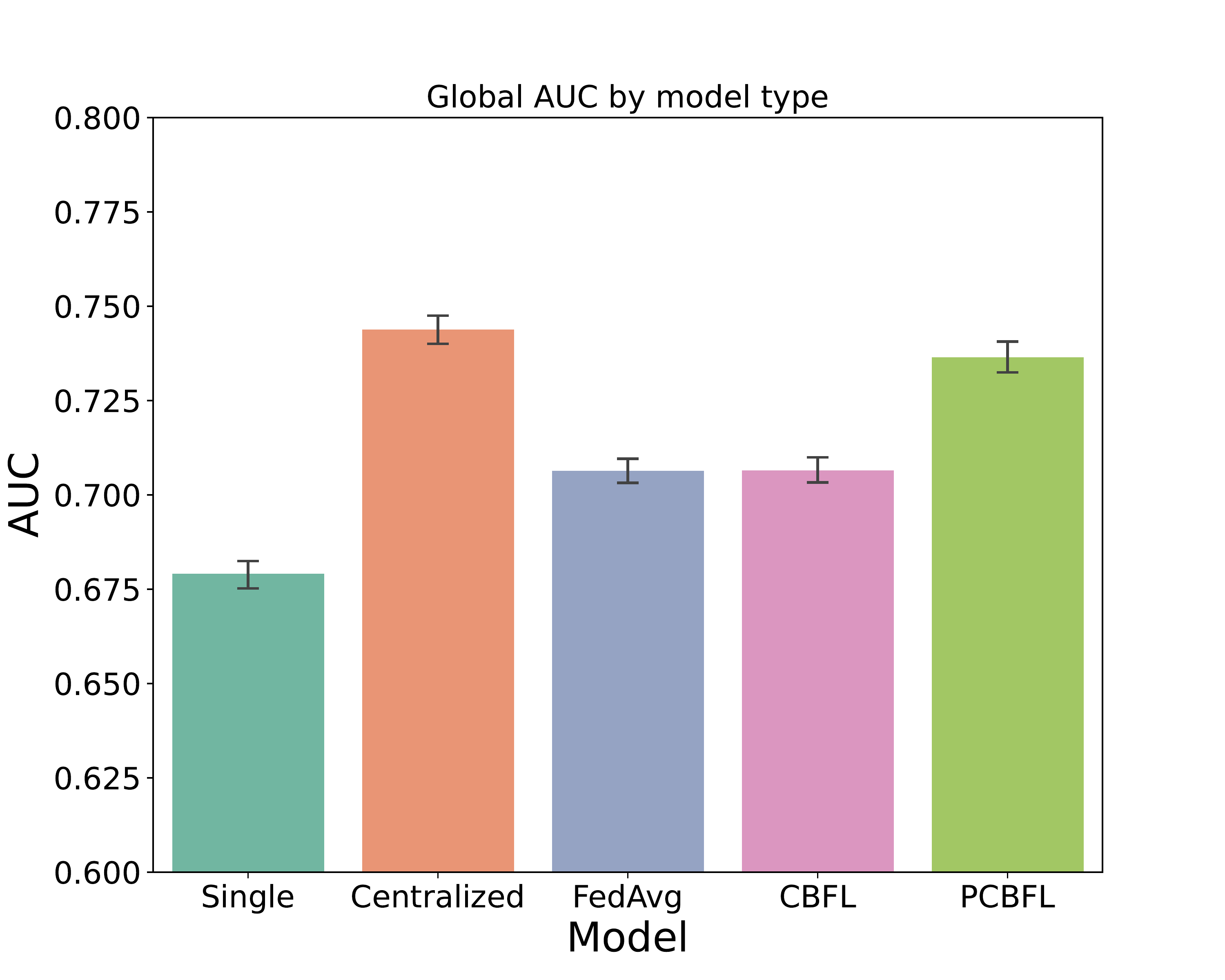}
}\hfill 
\subfigure{%
\label{fig:pic2}
\includegraphics[width=0.49\textwidth]{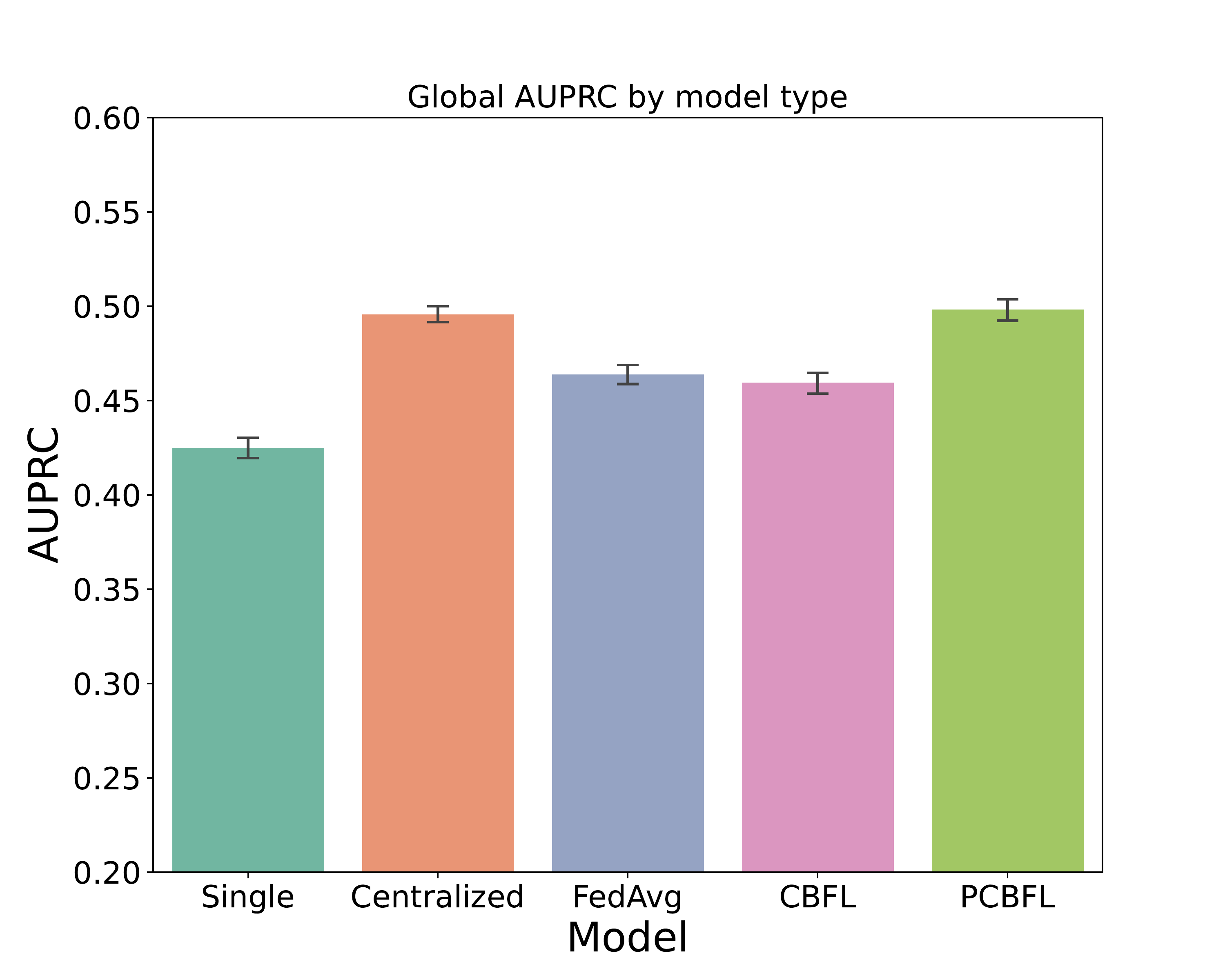}
}
}
\end{figure}
\vspace*{-3mm}

\subsection{PCBFL enables better predictive performance at most sites} 
Figure 4 shows the number of sites where each model has the highest performance. We compared Single site, FedAvg, CBFL and PCBFL at 20 sites. Centralized training was not compared as there are no per-site results. PCBFL performs  best at 12 and 9 sites in terms of AUC and AUPRC, respectively. Figure 5 shows the AUC scores for each site and cluster (see Supplementary Figure 2 for AUPRC). PCBFL outperforms single site training at 16 and 18 sites, FedAvg at 14 and 14 sites, and CBFL at 13 and 13 sites for AUC and AUPRC, respectively.

\begin{figure}[!ht]
\floatconts
{fig:example2}
{\vspace*{-8mm}\caption{\textbf{Number of sites where model has highest} AUC (a) and AUPRC (b) for Single, Centralized, FedAvg, CBFL and PCBFL.}}
{%
\subfigure{%
\label{fig:pic1}
\includegraphics[width=0.49\textwidth]{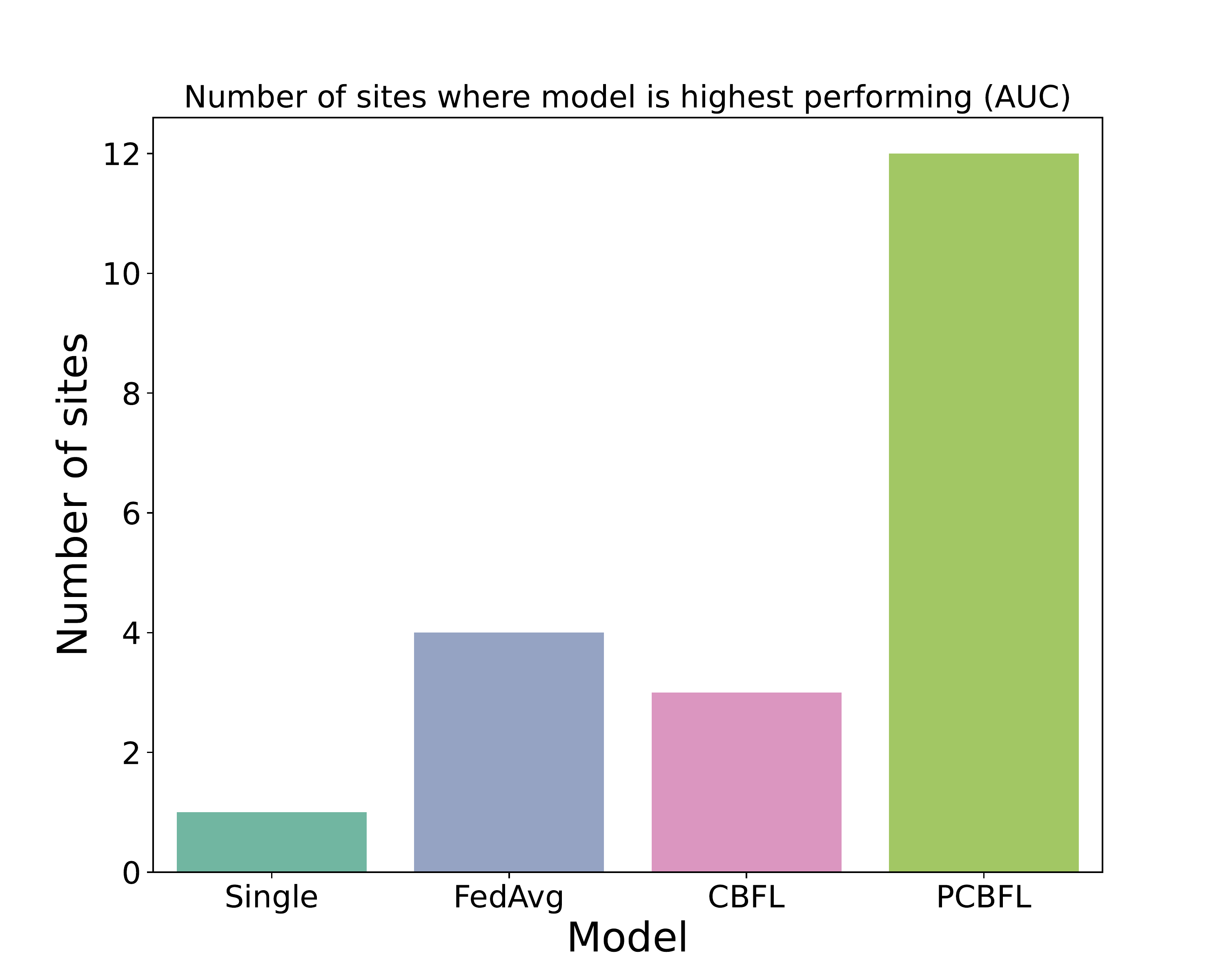}
}\hfill 
\subfigure{%
\label{fig:pic2}
\includegraphics[width=0.49\textwidth]{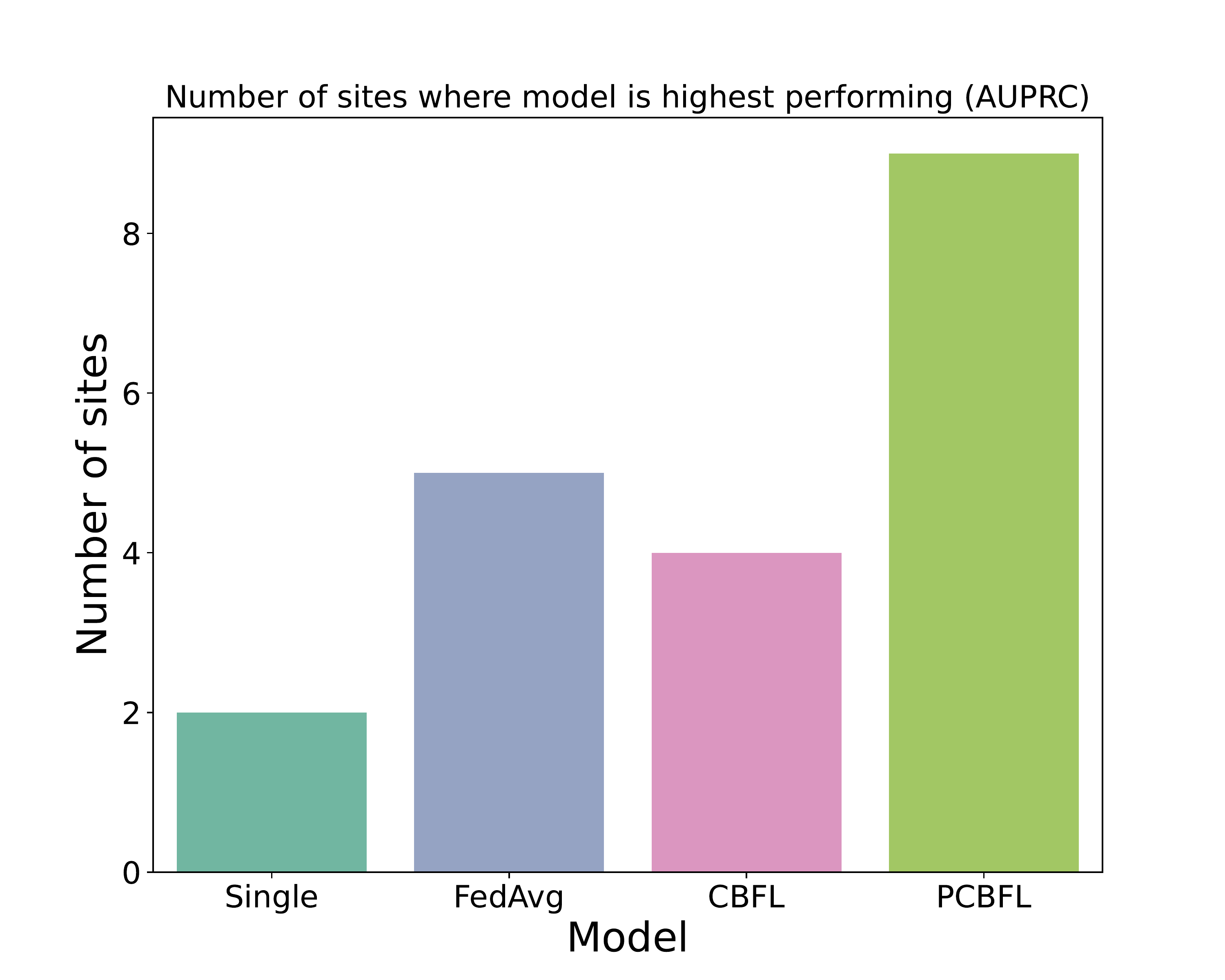}
}
}
\end{figure}

\begin{figure}[!ht]
\floatconts
{fig:example2}
{\vspace*{-8mm}\caption{\textbf{Model performance by site}, AUC. Results for Single, FedAvg, CBFL and PCBFL.}}
{%
\label{fig:pic1}
\includegraphics[width=1\textwidth]{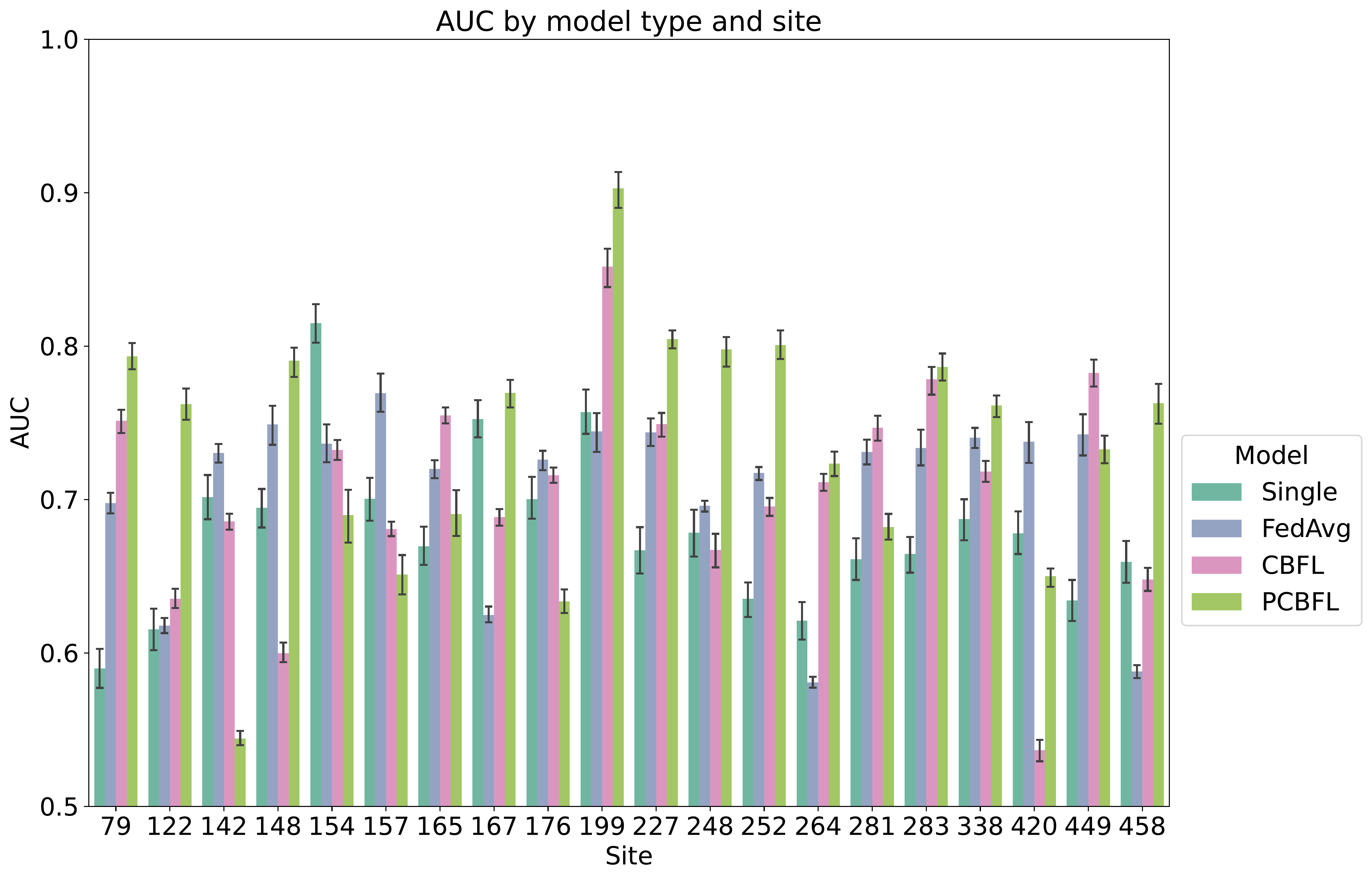}
}
\end{figure}

\section{Discussion} 
We present a new personalized FL framework based on privacy-preserving patient clustering (PCBFL). We show that this algorithm enables better performance in a downstream mortality prediction task across 20 ICU datasets when compared to traditional FL and existing Clustered FL techniques. Furthermore, per site analysis shows that PCBFL is most likely to achieve best performance at any given site. PCBFL also generates clinically meaningful clusters categorizing patients into low, medium, and high mortality risk groups based on physical exam, diagnosis, and medication values. Our results show that PCBFL can be used to implement personalized FL, addressing the challenges of non-IID EHR data and patient privacy by securely clustering patients and optimizing model performance on each cluster. The ability to generate clinically meaningful subgroups suggests PCBFL can be extended to other clinical use cases such as phenotyping, risk stratification, and advancing disease understanding. 

We demonstrated that PCBFL is able to generate clinically meaningful groups that were categorized mostly based on patient severity. This is in contrast to CBFL, which has been shown to cluster patients based on geographical distribution of the hospitals (\cite{Huang2019-pm}). PCBFL also enables clustering over a very large number of sites and patients and is able to do so without introducing error into the calculation. These findings suggest that PCBFL can be extended to support the discovery of novel subgroups without the need for a prediction task (\textit{e.g.,} unsupervised phenotyping, risk stratification, disease subtyping, treatment selection, and trial recruitment) (\cite{Robinson2012-mx}). Previous studies have shown success of clustering patients in centralized settings and we believe it can now be extended to the federated settings (\cite{Xu2020-ky,Zeng2021-yh}). Overall, our study highlights the potential of PCBFL as a powerful tool for collaborative analysis of healthcare data.

We also found that PCBFL demonstrates a meaningful improvement in global performance compared to traditional FL frameworks (FedAvg) and other personalized FL frameworks (CBFL). We believe this improvement can be attributed to our clustering technique, which is able to divide the federated datasets into more IID cohorts by using patient similarity scores between individual patients. This likely improves the performance of the downstream tasks by reducing the impact of site-specific biases and allowing the model to focus on cohort-specific features necessary for predictions. As a result, it has the potential to improve the generalizability of the models (\cite{Prayitno2021-gm,fallah2020personalized}). Moreover, we found that PCBFL has the best per-site performance compared to other methods, which increases motivation for sites to participate in federated learning (\cite{Cho2022-se}).

\paragraph{Limitations:} PCBFL has some limitations. First, it relies on sufficient sample sizes per cluster, which can be an issue in cases where sites have limited datasets. However, given the improved performance against single site training, in cases where sufficient samples are available, clustering should be preferred. Second, the secure clustering algorithm requires pairwise cosine similarity calculation across all sites. This results in additional communication costs as each pair of hospitals must use their own separate masking matrix and secret sharing protocols. A new secret sharing scheme with central server coordination can be developed that reuses secret shares across multiple calculations, thus reducing the communication cost. In addition, PCBFL requires training of two deep learning models and a clustering algorithm. As such, the communication cost is higher than traditional FL frameworks. However, we feel the trade-off between communication cost and improved model performance is acceptable in the context of healthcare, where higher accuracy is preferred over compute power. Finally, this analysis is limited to eICU dataset and a mortality prediction task. PCBFL should be assessed on a range of clinical prediction tasks and datasets to fully evaluate its performance.

\section{Conclusion} 
We present a new personalized FL framework based on a novel privacy-preserving patient clustering algorithm (PCBFL) that addresses the challenge of non-IID data and patient privacy in federated settings. Our study demonstrates that PCBFL enables better model performance than existing methods in a mortality prediction task. We showed that the clustering technique used by PCBFL divides the federated datasets into clinically meaningful cohorts suggesting it can be extended to other phenotyping tasks. These findings highlight the potential of PCBFL as a powerful tool for collaborative analysis of healthcare data. In future work, we plan to explore the generalizability of PCBFL to other healthcare datasets and domains.

\bibliography{main.bbl} 
\appendix
\section*{Appendix}

\section{Selecting number of clusters}
\subsection{Within-Cluster-Sum-Of-Squares}
Within-Cluster-Sum-Of-Squares (WCSS) is a metric used in clustering to determine the optimal number of clusters. It calculates the total sum of the squared distances between each data point within a cluster and the center of that cluster (centroid). A smaller WCSS implies that the data points are more compact, indicating tighter clustering of similar points. It is defined as:
    \begin{equation} \label{eq1}
        \text{WCSS}:= \sum_{k=1}^{K}\sum_{i=1}^{n_k} (x_{ki} - \mu_{k})^2
    \end{equation}
where cluster $k$ ranges from $1..K$m data points, $x_{ki}$ in cluster $k$ range from $1..n_k$, and $\mu_k$ is the mean of the points in cluster $k$ \emph{i.e.,} $\frac{1}{n_k}\sum_{i=1}^{n_k}x_{ki}$
\subsection{Elbow plot}
\newcounter{suppfigure}
\renewcommand{\thesuppfigure}{S\arabic{suppfigure}}
\begin{figure}[!ht]
\centering 
\includegraphics[width=9cm, height=7cm]{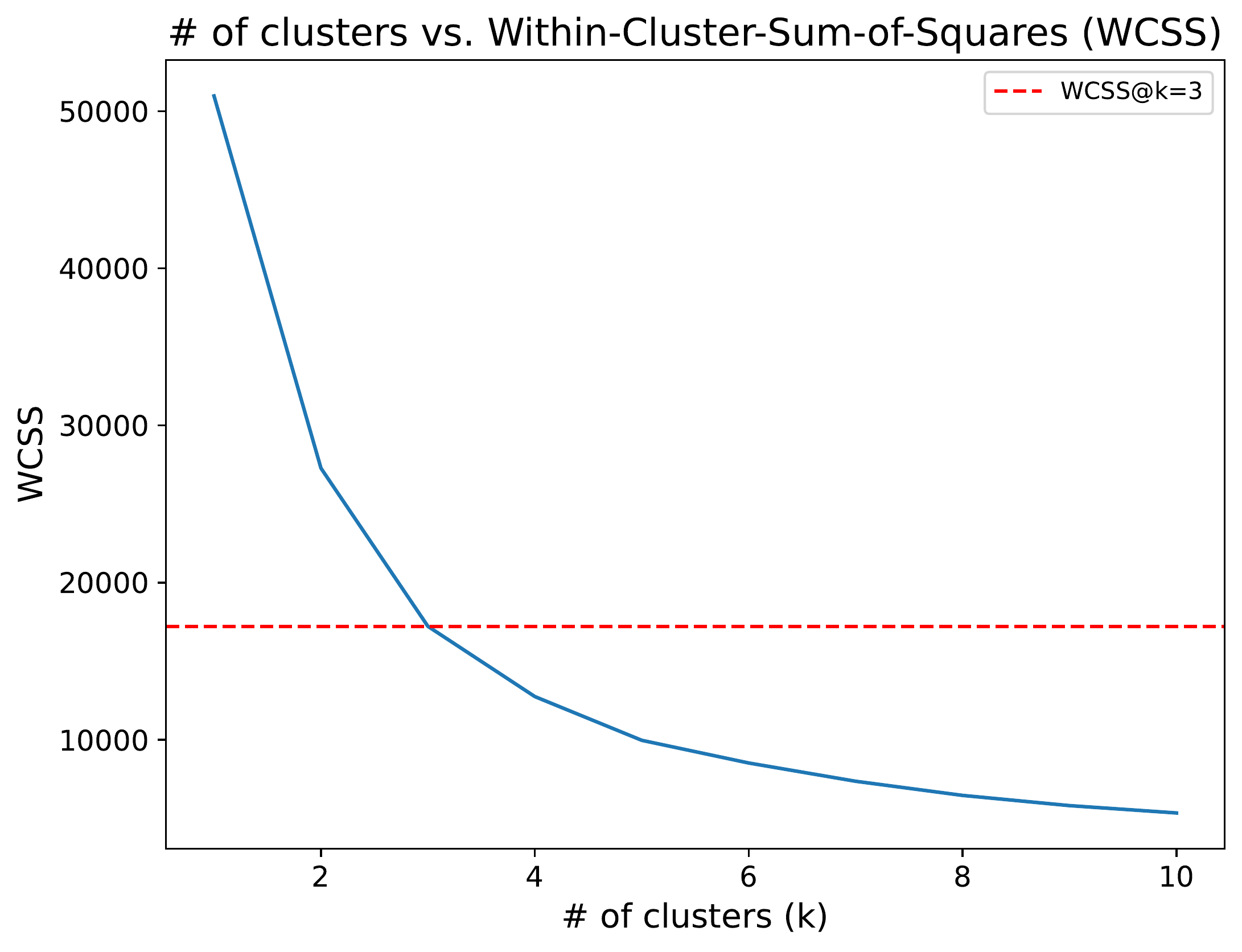} 
\vspace*{-6mm}
\stepcounter{suppfigure}
\caption*{\textbf{Supplementary Figure \thesuppfigure.} Number of clusters vs. Within-Cluster-Sum-Of-Squares (WCSS). Dashed red-line indicates WCSS at k=3.}
\label{sfig:elbow} 
\end{figure} 

\section{Calculating overall performance}
\subsection{Single site and FedAvg}
To calculate the overall AUC and AUPRC for Single Site and FedAvg (Global R), we use a simple average (or uniform weighted average). This is because each site trains only one model and has the same number of patients:
    \begin{equation} \label{eq1}
        \text{Global R}= \frac{1}{N}\sum_{c=1}^{C}R_{c}n_c
    \end{equation}
where $R_c$ is the result (AUC or AUPRC) for site $c$, $n_c$ is the sample size for site $c$, and $N$ is the total number of samples across all sites.
\subsection{PCFBL and CBFL}
To calculate the overall AUC and AUPRC (Global R) for CBFL and PCBFL, we use a weighted average of results that takes into account the number of patients each site contributes to a cluster. This is necessary as each site trains 3 separate models (1 per cluster) and their sample size contribution to a given cluster varies:
    \begin{equation} \label{eq1}
        \text{Global R}= \frac{1}{N}\sum_{c=1}^{C}\sum_{k=1}^{K}R_{ck}n_{ck}
    \end{equation}
where $R_{ck}$ is the result (AUC or AUPRC) for site $c$ and cluster $k$, $n_{ck}$ is the sample size for site $c$ and cluster $k$, and $N$ is the total number of samples across all sites.

\section{Cluster regional distribution}
\begin{table}[H]
\centering
\begin{tabular}{|llll|}
\hline
\textbf{Region}            & \textbf{Cluster} & \textbf{PCBFL} & \textbf{CBFL} \\ \hline
\multirow{3}{*}{\textbf{Midwest}}   & Low              & 36.7           & 46.4          \\
                           & Medium           & 19.85          & 48.25         \\
                           & High             & 43.45          & 5.35          \\ \hline
\multirow{3}{*}{\textbf{Northeast}} & Low              & 44.2           & 36            \\
                           & Medium           & 15.6           & 61.2          \\
                           & High             & 40.2           & 2.8           \\ \hline
\multirow{3}{*}{\textbf{South}}     & Low              & 39.53          & 18.27         \\
                           & Medium           & 16.47          & 81.47         \\
                           & High             & 44             & 0.27          \\ \hline
\multirow{3}{*}{\textbf{West}}      & Low              & 51.87          & 24.6         \\
                           & Medium           & 22.6           & 10.73         \\
                           & High             & 25.53          & 64.67         \\ \hline
\end{tabular}
\vspace{1mm}
  \label{tab:example}  \\
  \textbf{Supp. Table 1.} Cluster distribution of each region for PCBFL and CBFL. All values expressed as a \% of patients in the region
\end{table}
\section{Per site results: AUPRC}
\begin{figure}[H]
{%
\includegraphics[width=1\textwidth]{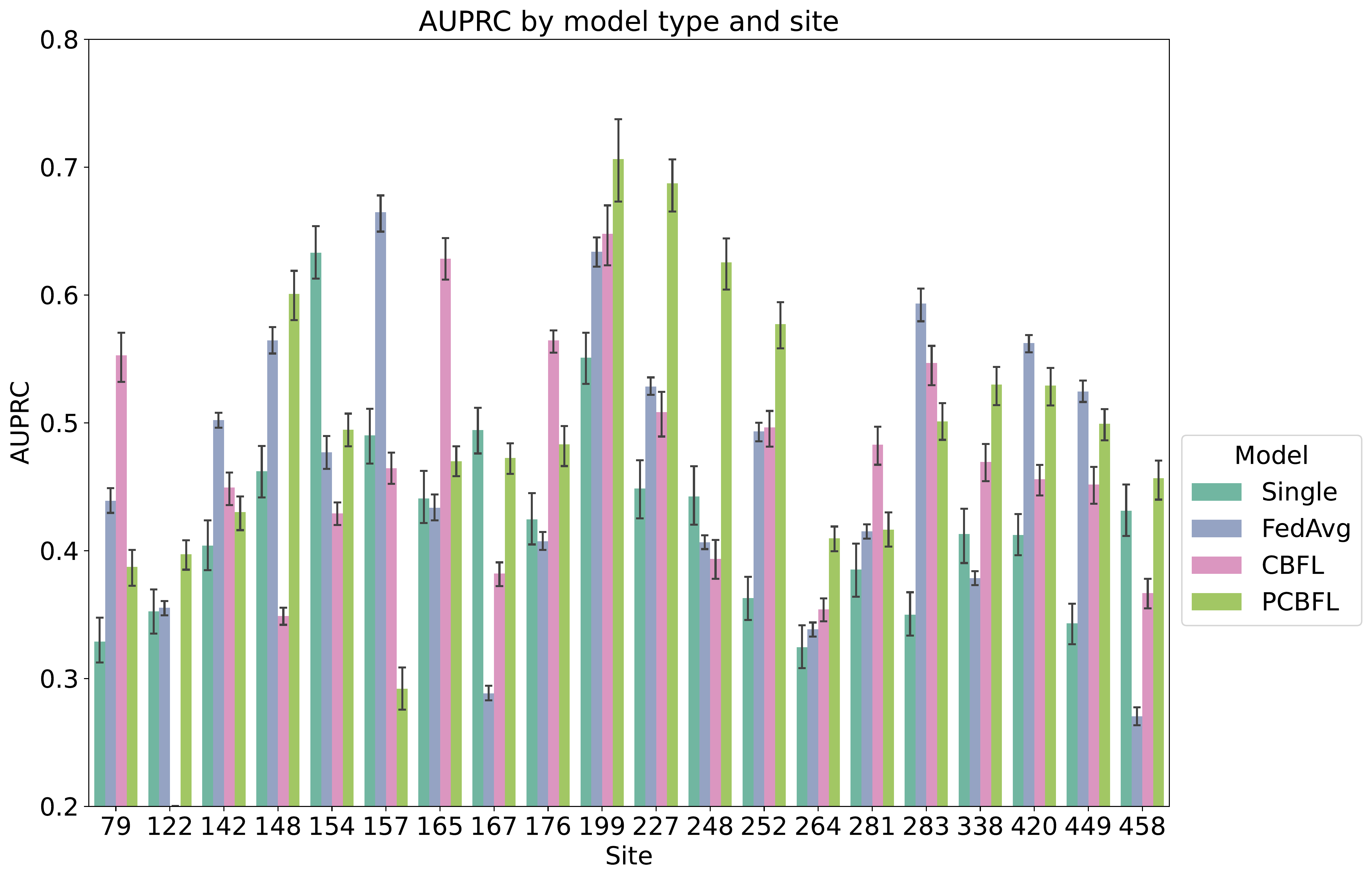}
\caption*{\textbf{Supplementary Figure \thesuppfigure.} Model performance by site, AUPRC. Results for Single, FedAvg, CBFL and PCBFL. \stepcounter{suppfigure}}
}
\end{figure}
\clearpage
\section{Pseudocode}
\begin{algorithm}
\caption{Creating patient embeddings}
\label{alg:Creating patient embeddings}
\DontPrintSemicolon
\KwIn{A set of patients across multiple sites, each associated with domains: observation, diagnosis, drug.}
\KwOut{The patient embedding matrix $\mathbf{E}$.}

\For{each $t$ in domains observation, diagnosis, drug}{
    Initialize weights, $w_{ae}$, of autoencoder model $f_{ae}$\;
    \For{each round $i$ from $1$ to $20$}{
        \For{each site $c$ from $1$ to $C$ in \textbf{parallel}}{
            Train $f_{ae}$ for 20 epochs to obtain $w^{c}_{i,ae}$\;
            \tcp{$c$ is the site index and $i$ is the training round index}
            \textbf{return} $w^{c}_{i,ae}$ to server\;
        }
        Server updates weights using $w_{i+1,ae} \gets \sum^{C}_{c=1}\frac{n^c}{N}w^{c}_{i,ae}$\;
        \tcp{$n^c$ is number of patients at site $c$, $N$ is total number of patients}
    }
    set final model $f^{g}_{ae} \gets f_{ae}$\;
}
\For{each site $c$ from $1$ to $C$ in \textbf{parallel}}{
    Create embedding matrix $\mathbf{E}^{|P_c|\times D}$\;
    \tcp{$|P_c|$ is the number of patients at site $c$ and $D$ is the embedding dimension}
    \For{each patient $p$ in patient set, $P_c$}{
        Create empty patient embedding vector, $\mathbf{E}^{1 \times D}_p$\;
        \For{each $t$ in domains observation, diagnosis, drug}{
            Create domain patient embedding, $\mathbf{E}_{p_t} = f^{g}_{ae}(p_t)$\;
            Concatenate $\mathbf{E}_p$ with $\mathbf{E}_{p_t}$\;
        }
        $\mathbf{E}[p] \gets \mathbf{E}_p$\;
    }
    Return  $\mathbf{E}$\;
}
\end{algorithm}

\begin{algorithm}
\caption{Estimating patient similarity}
\label{alg:Estimating patient similarity}
\DontPrintSemicolon
\KwIn{Patient embeddings across multiple sites, $\mathbf{E}_p$.}
\KwOut{Global similarity matrix $S$.}

Initialize global similarity matrix, $S$, as a $P \times P$ zero matrix, where $P$ is the total number of patients\;
\For{every unique pair of sites $c$, $d$ $\in$ sites C in \textbf{parallel}}{
    Construct invertible matrix, $M_{f\times f}$ using Reed-Hoffman encoding with degree $f > D$ \emph{i.e.,} degree greater than embedding dimension\;
    Send $M$ to $c$ and $M^{-1}$ to site $d$\;
    \For{site $c$}{
        Set matrix $A$ to hold all patient embeddings $\mathbf{E}_p$ within site $c$\;
        Calculate $A_1 = A \times M_{left}$, $A_2 = A \times M_{right}$\;
        Send $A_1$ to the server\;
    }
    \For{site $d$}{
        Set matrix $B$ to hold all patient embeddings $\mathbf{E}_p$ within site $d$\;
        Calculate $B_1 = B \times M^{-1}_{top}$, $B_2 = B \times M^{-1}_{bottom}$\;
        Send $B_2$ to the server\;
    }
    \For{site $c$}{
        Receive $B_2$ from the server\;
        Calculate $V_2 = A_2B_2$\;
        Send $V_2$ to the server\;
    }
    \For{site $d$}{
        Receive $A_1$ from the server\;
        Calculate $V_1 = A_1B_1$ \;
        Send $V_1$ to the server\;
    }
    Server completes $V$=$V_1$+$V_2$\;
    \For{ $\forall$ patient $i \in c$}{
        \For{ $\forall$ patient $j \in d$}{
            Update similarity matrix $S$ with the corresponding value in $V$: $S_{c_i,d_j}$, $S_{d_j,c_i}$ $\gets$ $V_{i,j}$\;
        }
    }
}
Return Global similarity matrix $S$\;
\end{algorithm}

\begin{algorithm}
\caption{Clustering patients}
\label{alg:Clustering patients}
\DontPrintSemicolon
\KwIn{Global similarity matrix $S$.}
\KwOut{Patient clusters for each site $c$ in site $C$.}

Initialize list to hold within-cluster sum of squares (WCSS) for each $k$, denoted as WCSS[]\;
\For{each $k$ $\in$ {1, 2, ..., 10}}{
    Apply spectral clustering to global similarity matrix $S$ with $k$ clusters\;
    Compute WCSS for current $k$, denoted as WCSS$_k$\;
    Store WCSS$_k$ in WCSS[$k$]\;
}
Determine optimal number of clusters, $k_{opt}$, using elbow method on WCSS\;
\For{each $c$ $\in$ site $C$}{
    Apply spectral clustering on $S$ using $k_{opt}$ clusters\;
    Return patient clusters for site $c$\;
}
\end{algorithm}

\begin{algorithm}
\caption{Predicting Outcomes}
\label{alg:Predicting Outcomes}
\DontPrintSemicolon
\KwIn{Clusters $K$}
\KwOut{Global AUC, Global AUPRC}
\For{each cluster $k$ $\in K$}{
    Initialize weights $w_{pk}$ of prediction model, $f_{pk}$\;
    \For{each round $i$ in 1 to 20}{
        \For{each site $c \in$ sites $C$}{
            Train $f_{pk}$ for 20 epochs to obtain $w^{c}_{i,pk}$\tcp*{$c$ is the site index and $i$ is the training round index}
            \Return $w^{c}_{i,pk}$ to server\;
        }
        Server updates weights $w_{i+1,pk}$ $\gets$ $\sum^{C}_{c=1}\sum^{K}_{k=1}\frac{n^{ck}}{N^{k}}w^{c}_{i,pk}$\tcp*{$n^{ck}$ is number of patients at site $c$ in cluster $k$, $N^{k}$ is total number of patients in cluster $k$}
        \For{each site $c \in C$}{
            Measure AUC$_{c,k}$ and AUPRC$_{c,k}$ in test set and send results\;
        }
    }
    Global AUC $\gets$ $\sum^{C}_{c=1}\sum^{K}_{k=1}\frac{n^{ck}}{N^{k}}$ AUC$_{c,k}$\;
    Global AUPRC $\gets$ $\sum^{C}_{c=1}\sum^{K}_{k=1}\frac{n^{ck}}{N^{k}}$ AUPRC$_{c,k}$\;
}
\Return Global AUC, Global AUPRC\;
\end{algorithm}
\clearpage
\section{Feature distributions by cluster}
\href{https://github.com/G2Lab/pcfbl/blob/main/suppl_tables/supp_tables_2_7.pdf}{Supplementary tables 2-7 can be found here.}

\end{document}